\pdfoutput=1

\documentclass[dvipsnames]{article}
\usepackage{arxiv}

\usepackage{amsmath}
\usepackage[authoryear]{natbib}
\usepackage[hidelinks]{hyperref}

\setcitestyle{authoryearh} %

\usepackage{amsthm,amsmath}
\usepackage[utf8]{inputenc} %

\usepackage{booktabs}
\usepackage{enumitem}
\usepackage{cleveref}
\usepackage{multirow}
\usepackage{subcaption}
\usepackage{amssymb}
\usepackage{xurl}
\usepackage{graphicx}

\newcommand{\data}{\mathbf{x}}
\newcommand{\Data}{\mathcal{X}}
\newcommand{\exprresultscaption}{Experimental results of the benchmarking ML models}

\newcommand{\shortheadtitle}{Ensemble Learning based Anomaly Detection for IoT Cybersecurity}

\begin{document}

\newcommand{\PaperTitle}{Ensemble Learning based Anomaly Detection for IoT Cybersecurity via Bayesian Hyperparameters Sensitivity Analysis}

\title{\PaperTitle}

\author{
  Tin Lai$^{\dagger,1}$
  \And
  Farnaz Farid$^{\mathsection,2}$
  \And
  Abubakar Bello$^{\mathsection,3}$
  \And
  Fariza Sabrina$^{\ddagger,4}$
  \AND
  \normalfont $^{\dagger}$School of Computer Science\\
  The University of Sydney\\
  Australia \\
  \And
  \normalfont $^{\mathsection}$School of Social Sciences\\
  Western Sydney University\\
  Australia \\
  \And
  \normalfont $^{\ddagger}$School of Engineering and Technology\\
  Central Queensland University\\
  Australia \\
}

\renewcommand\footnotemark{}
\thanks{
  \noindent
  $^{1}$\texttt{tin.lai@sydney.edu.au}~~
  \texttt{\{$^{2}$farnaz.farid,$^{3}$a.bello\}@westernsydney.edu.au}~~
  $^{4}$\texttt{f.sabrina@cqu.edu.au}
}

\maketitle

\begin{abstract}
  The Internet of Things (IoT) integrates more than billions of intelligent devices over the globe with the capability of communicating with other connected devices with little to no human intervention. IoT enables data aggregation and analysis on a large scale to improve life quality in many domains. In particular, data collected by IoT contain a tremendous amount of information for anomaly detection. The heterogeneous nature of IoT is both a challenge and an opportunity for cybersecurity. Traditional approaches in cybersecurity monitoring often require different kinds of data pre-processing and handling for various data types,  which might be problematic for datasets that contain heterogeneous features. However, heterogeneous types of network devices can often capture a more diverse set of signals than a single type of device readings, which is particularly useful for anomaly detection. In this paper, we present a comprehensive study on using ensemble machine learning methods for enhancing IoT cybersecurity via anomaly detection. Rather than using one single machine learning model, ensemble learning combines the predictive power from multiple models, enhancing their predictive accuracy in heterogeneous datasets rather than using one single machine learning model. We propose a unified framework with ensemble learning that utilises Bayesian hyperparameter optimisation to adapt to a network environment that contains multiple IoT sensor readings. Experimentally, we illustrate their high predictive power when compared to traditional methods.
\end{abstract}


\section{Introduction}

We live in a modern era where internet-connected devices are ubiquitous, and cybersecurity threats persist everywhere. The connected devices, often known as the Internet of Things (IoT), refer to all electronic devices connected to the internet or other devices~\citep{lee2015_InteThin}. IoT devices are capable of transmitting and collecting data for various tasks. For example, nowadays, household appliances, cars, tools, or personal devices can sense, process, and connect to the internet~\citep{gonzalez-zamar2020_IoTTech}. The development of technologies enables the usage of IoT in urban environments, creating smarter cities~\citep{okano2017_IOTIndu}. Devices can now communicate with each other, such as household appliances, consumer devices, sensors, or even industrial controls \citep{okano2017_IOTIndu}.
IoT products resemble a powerful approach for increasing connectivity between devices. Altogether it represents interconnected devices and services that can communicate and share data and information among various domains and applications. IoT lead systems can transform our lives by making intelligent decisions to improve the quality of daily tasks. For example, seamless home integration, smart city infrastructures, and transportation are relevant sectors in which IoT devices can be used.
The high availability of data generated by IoT devices enables researchers to perform extensive scientific analysis on data mining and extracting the underlying relationship that is otherwise hard to discover~\citep{kaur2018_AnalScie}.
However, like many other new innovations, it comes with specific security risks ~\citep{wheelus2020_IoTNetw}. By extending internet connections to everyday devices, these threats have expanded from our homes to workplaces, healthcare and other facilities.
Many security risks exist that enable various electronic devices to exhibit the capability to record and send received signals to remote locations over the internet~\citep{hassija2019_SurvIoT}.
However, this unified capability also opens a powerful opportunity in the application of anomaly detection.

Cybersecurity threat modelling and detection via anomaly detection is a multidisciplinary problem with applications in various domains. It involves identifying unusual or unexpected observations within the captured data due to uncommon values or data sequence~\citep{chandola2009_AnomDete}.
Anomaly detection finds patterns in data that do not conform to expected behaviour, and these non-conforming patterns are referred to as \emph{anomalies}.
Traditional anomaly detection applications are typically used extensively in intrusion detection, spotting malicious activities and even in safety-critical systems like military surveillance.
The usage is applicable in numerous domains, and anomalies in data often translate to critical actionable information.
Anomalies might be induced due to various reasons like malicious activity, credit card fraud, cyber-intrusion, breakdown of a system, or malfunction of some components.
Therefore, the ability to detect such events provides us with an opportunity to react to such an anomaly event.

Moreover, anomalies in data often imply critical and actionable information essential to implement a secure network system.
Anomaly detection had been successfully used to identify thief~\citep{aleskerov1997_CardNeur}, presence of tumours in medical settings~\citep{spence2001_DeteSynt}, compromised computer sending sensitive data to remote location~\citep{kumar2005_ParaDist}, or even signifying faulty component within space craft~\citep{fujimaki2005_ApprSpac}.
Anomaly detection enhances cybersecurity's resilience and robustness in mission-critical applications.

The abundance of sensor data enables a new opportunity for vendors and device owners to utilise the sensor data for continuous system monitoring.
Using a machine-intelligent approach to tackle cybersecurity via automatic detection of anomaly events can better utilise the available information.
The use of machine learning (ML) in IoT devices is still relatively new compared to other domains~\citep {cook2020_AnomDete}.
In contrast to typical datasets in traditional ML, most data sources collected in IoT devices are time-series because of the continuous monitoring nature of the sensory devices.
Most \emph{contextual anomalies} are observations that deviate from the expected patterns within the time-sires~\citep{chandola2009_AnomDete}.
The recent advancement of Deep Learning (DL) also enables us to utilise them as a universal function approximator to extract patterns that are had to hand-craft automatically.
For example, DL can detect suspicious activities in a practical real-time system for intrusion detection~\citep{alghamdi2021_DeepIntr}.

In this study, we improve the robustness of the existing state-of-the-art ensemble machine learning model in cybersecurity via a systematic hyperparameter optimisation process. We comprehensively investigate the predictive performance of traditional machine learning and ensembles of ML models in a wide range of datasets. In particular, machine learning models tend to be sensitive to the choice of hyperparameters, which can significantly affect the model's predictive accuracy.
Our contributions are summarised as follows.
\begin{enumerate}[label=\roman*)]
  \item We present an in-depth empirical study on ensemble models' predictive performance for IoT Cybersecurity, which combines multiple weak predictors into ensembles of models with a much higher predictive capability.
  \item We investigate the choice of hyperparameter via parametric sensitivity analysis and present the important set of hyperparameters for each type of ensemble model.
  \item We propose a Bayesian-based framework for training ensemble models that utilise Bayesian Optimisation techniques for automatic search for the best set of hyperparameters via optimising a surrogate model.
  \item Finally, we summarise the essential network features for detecting network cybersecurity threats on IoT devices.
\end{enumerate}
Experimentally, we demonstrate the Bayesian-based framework's effectiveness, which can improve the F1 score of state-the-art ensemble models by 10\% to 30\% for models with and without hyperparameter optimisation.

\section{Related works}

IoT systems have benefited the development and data usage across numerous domains.
However, several weaknesses exist across IoT systems, for example, vulnerability, security, device disruption, data theft, interruption, and Main-in-The-Middle attacks~\citep{gou2013_ConsStra}.
In addition to the research and development of communication techniques, a large amount of work within IoT involves imposing security measures~\citep {hassan2019_CurrRese} and bridging with the domain of cyber computing.
For example, it is essential to strategise a security system~\citep{hassija2019_SurvIoT-1} for minimising the potential vulnerability and composing network architectures that address security threats~\citep{hwang2015_IotSecu}.
Most recent researches focus on challenges and status within IoT security measure~\citep{mahmoud2015_InteThin}, IoT exploitations~\citep{neshenko2019_DemyIoT}, possible approaches to enhance security~\citep{riahi2013_SystAppr} and possible resolution by employing a deep learning approach~\citep{al-garadi2020_SurvMach}.
Innovative smart city applications connect enormous IoT devices to real-world objects spanning large distances and can often provide essential benefits to urban life~\citep{borgia2014_InteThin}.
Moreover, the massive number of IoT devices contain heterogeneous services and protocols, which leads to the complexity of managing numerous devices across networks.
Therefore, integrating devices without proper management might introduce serious cybersecurity threats and vulnerabilities for malicious actors to attack and extract daily activities and information about the average citizen's life.

Machine Learning (ML) is a powerful approach to enhancing the detection of IoT cyberattacks and malicious events. It is an integral part of extracting patterns within data~\citep{mahesh2020_MachLear} for providing a data-driven approach to creating predictive models~\citep{carbonell1983_OverMach}.
There have been lots of recent development in artificial intelligence in various domains; for example, in computer vision~\citep{geraldes2019_UAVbSitu} and robotics~\citep{lai2021_PlanLear} domains, for problems such as classification~\citep{kotsiantis2007_SupeMach}, extracting patterns from time series data~\citep{wang2021_LearNonS}, for processing sequential data~\citep{dietterich2002_MachLear}, and in medical domain as a predictive tool~\citep{sensorsMLforDiabetes}.
A smart city calls for the use of more innovative systems, Big Data analytics and the use of pattern recognition for detecting trends that are abnormal from previous observations. For example, we can associate the current observation of some sensor data against previously observed data to assess the degree of deviations from the expected value defined by the anomaly detection model. There are a variety of methods of generating anomaly scores that are unique to each detection algorithm and model. Anomaly scoring is helpful in the identification and management of outliers when performing analytical tasks such as predictive analytics.

Machine Learning allows for utilising sensor data from IoT devices to improve their associate security~\citep{xiao2018_IoTSecu}.
Using pattern matching and computing statistical inference can help to address some of the ongoing IoT security challenges~\citep{zhang2014_IoTSecu}.
The collected IoT sensor data typically contain temporal components as they are some time series describing monitoring system attributes.
Anomaly detection in time series is a well-known problem that has been tackled with a long history~\citep{hawkins1980_IdenOutl,abraham1989_OutlDete}.
We refer interested readers to existing surveys like~\citep{markou2003_NoveDete} and~\citep{zhang2010_OutlDete} for a comprehensive overview of the various statistical approaches and network models employed in time-series data.

More recently, deep learning approaches have also been employed in the space of time-series anomaly detection, with some focus on applications in Industrial IoT. Most current approaches involve some statistical inference for utilising historical data for modelling the expected system behaviour. It involves pattern matching of anomaly events in a supervised-learning setting with known anomalous characteristics or distance-based approaches that detect anomalies by detecting outliers. Most existing techniques can be grouped into the following categories:
\begin{enumerate}[label=\roman*)]
\item  Statistical and Probabilistic approach---utilise historical data to model the expected behaviour of a system.
New observations are compared against the current model for the system of interest.When the observation does not agree with the model, the observation is registered as an anomaly.
\item Predictive model---where a regression model is used to predict potential future values, and when the predicted value does not agree with the observed values, they are flagged as an anomaly.
\item Clustering-based methods--- project the data into a high-dimensional space and uses the density of the resulting clusters to determine outliers.
Significant clusters are typically indicated as regular observations, while outliers are identified as anomalies.
\item Pattern matching---directly models the time series in a supervised setting with known characteristics for anomalous data.

\end{enumerate}

New observations are compared against a database of labelled anomaly events and flags observations substantially different from data within the database as an anomaly.
Recent advancements in big data have enabled supervised learning methods in IoT settings due to the availability of sufficiently large IoT datasets.
The \texttt{IoT-23}~\citep{garcia2020_IoT2Labe} dataset is a recent network traffic dataset that consists of recorded data from multiple intelligent home IoT devices.
The recording devices include \emph{Amazon Echo}, \emph{Phillips HUE}, and \emph{Somfy Door Lock}.
The dataset contains real and labelled IoT malware infections and benign traffic.
The mass adoption of IoT devices enables researchers to access numerous datasets.
For example, MQTTset~\citep{vaccari2020_MQTTNew} is another dataset that recorded network traffics under MQTT protocol.
The availability of IoT deployment has increased the integration pace and extended the internet to multiple physical devices in our physical world.
This paper comprehensively studies the current state-of-the-art ensemble models on IoT anomaly detection tasks on multiple cybersecurity datasets.

\section{Methodology}

We propose a unified framework for evaluating Machine Learning models for use in an internet-connected environment via a monitoring network for detecting cybersecurity related anomaly events. The proposed methodology implements a detector for anomalies. Systematically, we analyse the overall performance of various machine learning models and their respective model architectures.

To identify IoT machine learning models that perform well across a wide range of domains, we empirically evaluate the performance of various baseline models by optimising their hyperparameters and their sensitivity toward the choice of parameters.
Our framework uses a Bayesian approach in searching for the most optimal set of hyperparameters.
Bayesian optimisation is a sequential design strategy that aims to optimise some objective value, which corresponds to the anomaly detection performance in our scenario.
Then, we present a set of ensemble models capable of combining the strength of multiple weaker predictive models into one that achieves superior predictive performance.
This section describes the systematic procedure of data preprocessing, data cleaning, feature engineering, and the various models used in this study.
Then, in~\cref{sec:model-results}, we present the empirical results obtained from our benchmark models, followed by a discussion of model sensitivity and important parameters in~\cref{sec:hyperparams}

\subsection{Data Preprocessing}

We first standardise each input feature in our evaluation by removing the mean and scaling to unit variance from the dataset $\Data$.
Let $\data_i \in \Data$ denote the $i^\text{th}$ datapoint, where we will use $\data_{i,j}$ to index into the $j^\text{th}$ feature of the datapoint.
We perform the standardisation by transforming each feature $\data_{i,j}$ to $\data'_{i,j}$ by
\begin{equation}
  \data'_{i,j} = \frac{\data_{i,j} - \overline{\data}_{j}}{\hat{\sigma}_j}
\end{equation}
where
\begin{equation}
  \overline{\data}_{j}=\frac{1}{|\Data|} \sum^{|\Data|}_{i=1} \data_{i,j},
\end{equation}
\begin{equation}
  \hat{\sigma}_j = \sqrt{\frac{1}{|\Data|} \sum^{|\Data|}_{i=1} (\data_{i,j} - \overline{\data}_{j}) }
    ,
\end{equation}
and $|\Data|$ denote the cardinality of $\Data$.
In addition, we compute pairwise Pearson correlation coefficient for the input features, where the correlation coefficient $r_{j,k}$ for the $j^\text{th}$ and $k^\text{th}$ feature is given by
\begin{equation}
    r_{j,k} =
    \frac{\sum^{|\Data|}_{i=1} (x_{i,j} - \overline{x}_j)(x_{i,k} - \overline{x}_k)}
    {\sqrt{
        \sum^{|\Data|}_{i=1} (x_{i,j} - \overline{x}_j)^2
        \sum^{|\Data|}_{i=1}(x_{i,k} - \overline{x}_k)^2}
    }
\end{equation}
which gives us the measure of linear correlation between the two features.
If $r_{j,k} > \delta$ for $\delta \in \mathbb{R}, 0 < \delta \le 1$, then we will remove the $k^\text{th}$ feature to help avoid overfitting.
The $\delta$ acts as a threshold to avoid highly correlated features, and in our experiments, we set $\delta = 0.7$.
We also convert all categorical features into one hot encoded feature, except for the IP address features.
It contains more than 5000 unique sparse categories in some scenarios, dramatically increasing the feature set size without providing meaningful predictive power.

\subsection{Anomaly Detection Dataset}

We evaluate each benchmark model against the following set of cybersecurity datasets to develop a model architecture that performs reasonably well across a range of IoT datastream domains.
In the following, we will briefly describe each dataset's content and the type of features the dataset contains.

The \texttt{IoTID20}~\citep{kim2019_IoTNetw} is an IoT intrusion dataset designed to be a comprehensive network dataset with flow-based features.
These flow-based features in this new botnet dataset help analyse flow-based intrusion detection systems, especially for monitoring anomalous activity across IoT networks.
The dataset contains captured attack packets from the smart home devices \emph{NUGU (NU 100)} and \emph{EZVIZ Wi-Fi Camera (C2C Mini O Plus 1080P)}, alongside some other laptops and smartphones within the same wireless network.
In particular, \texttt{IoTID20} contains 80 network features and detailed categories.
The dataset contains three variants for the labelled classes:
\begin{enumerate}
  \item \texttt{IoTID20 Binary} contains two possible classes. (i) \textit{Normal}: which indicates that there are no suspicious or malicious activities found within the connections, and (ii) \textit{Malicious}: which indicates malicious network traffics from infected devices.
  \item \texttt{IoTID20 Multi-Cat} contains multiple categories with five possible classes. (i) \textit{Normal}: represents the same set of flow data as the \texttt{Binary} dataset.
  The \textit{Malicious} is divided into individual classes, with
  (ii) \textit{DoS}: denoting network traffic that belongs to a Denial-of-Service attack;
  (iii) \textit{MITM ARP Spoofing}: denoting Man in the Middle attack by ARP spoofing;
  (iv) \textit{Mirai}: denoting traffics coming from devices that are infected by the Mirai malware, which will turn networked devices into remotely controlled bots;
  and
  (v) \textit{Scan}: which denote traffics scanning the IoT devices.
  \item \texttt{IoTID20 Multi-SubCat} divides the malicious traffic into sub-categories with nine possible classes. (i) \textit{Normal}: are the same as the previous variants;
  (ii) \textit{DOS-Synflooding}: denote Denial-of-Service attack by SYN flood;
  (iii) \textit{MITM ARP Spoofing}: denoting Man in the Middle attack by ARP spoofing;
  (iv) \textit{Mirai-Ackflooding}: denote traffics from Mirai Bot that is performing flooding with TCP ACK packets;
  (v) \textit{Mirai-HTTP flooding}: denote traffics from Mirai Bot that is performing flooding with HTTP requests;
  (vi) \textit{Mirai-Host Bruteforcing}: denote traffics from Mirai Bot that is performing brute force attack on a virtual host;
  (vii) \textit{Mirai-UDP flooding}: denote traffics from Mirai Bot that is performing flooding with User Datagram Protocol (UDP) packets;
  (viii) \textit{Scan Hostport}: denote traffics that is scanning the host for open ports; and
  (ix) \textit{Scan Port OS}: denotes traffic scanning the ports of the OS.
\end{enumerate}

The \texttt{IoT-23}~\citep{garcia2020_IoT2Labe} dataset is a recent network traffic dataset that consists of recorded data from multiple smart home IoT devices.
The recording devices include \emph{Amazon Echo}, \emph{Phillips HUE}, and \emph{Somfy Door Lock}.
The dataset contains real and labelled IoT malware infections and benign traffic.
In particular, \texttt{IoT-23} contains twenty-three captured scenarios, including twenty malicious network traffic and three benign traffic captures.
Moreover, the dataset recorded newer devices that the current cyber security systems have not interacted with before, which helps evaluate the current security measure against newer IoT devices.
In our evaluation of the candidate models, we clean the dataset by unifying the mismatched labels, loading up the first $1,000,000$ entries of each captured scenario, and removing classes containing less than five instances.
The dataset contains two variants for the labelled classes:
\begin{enumerate}
  \setcounter{enumi}{3}
  \item \texttt{IoT-23 Binary} contains two possible classes. (i) \textit{Benign}: which indicates that there are no suspicious or malicious activities found within the connections, and (ii) \textit{Malicious}: which indicates malicious network traffics from infected devices.
  \item \texttt{IoT-23 Multi-Cat} contains two possible classes. (i) \textit{Benign}: which indicates that there are no suspicious or malicious activities found within the connections, and (ii) \textit{Malicious}: which indicates malicious network traffics from infected devices.
\end{enumerate}
\Cref{table:iotid20} summarises the content of all the evaluating datasets.
The dataset content and associated labels are available to download in the linked address for reproducing our empirical study.
In the following, we will detail the models used in this study and our methodology for our Bayesian approach to optimising the hyperparameters.

\begin{table*}[tbh]
  \caption{A detailed overview of the distribution of classes within the evaluated dataset\label{table:iotid20}}
  \centering
\begin{tabular}{lcccccc}
\toprule
\multicolumn{1}{r}{\textbf{Dataset}}                                    && \textbf{Label}            & \textbf{} & \textbf{Train Set} & \textbf{Test Set} & \textbf{Label Instances} \\ \midrule
\multirow{3}{*}{\texttt{IoTID20}} & \multirow{3}{*}{\texttt{Binary}}    & Normal                    &           & 31,979             & 8,094             & 40,073                   \\
                                                                        && Anomaly                   &           & 468,353            & 116,989           & 585,342                  \\ \cmidrule(lr){4-7}
                                                                        &&                           & \textbf{Total}     & \textbf{500,332}            & \textbf{125,083}           & \textbf{625,415}                  \\ \midrule
\multirow{6}{*}{\texttt{IoTID20}} & \multirow{6}{*}{\texttt{Multi-Cat}}   & Normal                    &           & 31,979             & 8,094             & 40,073                   \\
                                                                        && DoS                       &           & 47,537             & 11,854            & 59,391                   \\
                                                                        && MITM ARP Spoofing         &           & 28,214             & 7,163             & 35,377                   \\
                                                                        && Mirai                     &           & 332,546            & 82,763            & 415,309                  \\
                                                                        && Scan                      &           & 60,056             & 15,209            & 75,265                   \\ \cmidrule(lr){4-7}
                                                                        &&                           & \textbf{Total}     & \textbf{500,332}            & \textbf{125,083}           & \textbf{625,415}                  \\\midrule
\multirow{10}{*}{\texttt{IoTID20}} & \multirow{10}{*}{\texttt{Multi-SubCat}}   & Normal                    &           & 31,979             & 8,094             & 40,073                   \\
                                                                        && DoS-Synflooding           &           & 47,537             & 11,854            & 59,391                   \\
                                                                        && MITM ARP Spoofing         &           & 28,214             & 7,163             & 35,377                   \\
                                                                        && Mirai-Ackflooding         &           & 44,117             & 11,007            & 55,124                   \\
                                                                        && Mirai-HTTP Flooding       &           & 44,643             & 11,175            & 55,818                   \\
                                                                        && Mirai-Host Bruteforcing   &           & 97,093             & 24,085            & 121,178                  \\
                                                                        && Mirai-UDP flooding        &           & 146,693            & 36,496            & 183,189                  \\
                                                                        && Scan Hostport             &           & 17,756             & 4,436             & 22,192                   \\
                                                                        && Scan Port OS              &           & 42,300             & 10,773            & 53,073                   \\ \cmidrule(lr){4-7}
                                                                        &&                           & \textbf{Total}     & \textbf{500,332}            & \textbf{125,083}           & \textbf{625,415}                  \\\midrule
\multirow{3}{*}{\texttt{IoT-23}} & \multirow{3}{*}{\texttt{Binary}}     & Benign                    &           & 1,462,947          & 366,180           & 1,829,127                \\
                                                                        && Malicious                 &           & 9,099,261          & 2,274,373         & 11,373,634               \\ \cmidrule(lr){4-7}
                                                                        &&                           & \textbf{Total}     & \textbf{10,562,208}         & \textbf{2,640,553}         & \textbf{13,202,761}               \\\midrule
\multirow{10}{*}{\texttt{IoT-23}} & \multirow{10}{*}{\texttt{Multi-Cat}}     & Benign                    &           & 1,462,947          & 366,180           & 1,829,127                \\
                                                                        && PartOfPortScan &           & 5,966,736          & 1,492,481         & 7,459,217                \\
                                                                        && Okiru                     &           & 2,102,304          & 523,948           & 2,626,252                \\
                                                                        && DDoS                      &           & 1,010,098          & 252,925           & 1,263,023                \\
                                                                        && Attack                    &           & 5,518              & 1,425             & 6,943                    \\
                                                                        && C\&C                       &           & 12,501             & 3,026             & 15,527                   \\
                                                                        && C\&C-HeartBeat             &           & 2,022              & 541               & 2,563                    \\
                                                                        && C\&C-FileDownload          &           & 63                 & 16                & 79                       \\
                                                                        && C\&C-Torii                 &           & 19                 & 11                & 30                       \\ \cmidrule(lr){4-7}
                                                                        &&                           & \textbf{Total}     & \textbf{10,562,208}         & \textbf{2,640,553}         & \textbf{13,202,761}               \\ \bottomrule
\end{tabular}
\end{table*}

\subsection{Bayesian Optimisation with Tree-structured Parzen Estimator:}
This study uses tree-Structured Parzen Estimator (TPE) to implement our automatic model training framework.
TPE is a sequential model-based Bayesian Optimisation (BO) approach, where it sequentially constructs models to approximate the performance of hyperparameters based on historical measurements.
Then, the algorithm updates its internal model and selects a new candidate of hyperparameters with a high potential for better performance.

Hyperparameter optimisation is formally defined as follows.
The performance of hyperparameters $x$ of some model $f$ can be measured by the mapping $f\colon\mathcal{X}\to\mathbb{R}$, where $\mathcal{X}$ denote the space of all possible hyperparameters and $\mathbb{R}$ is the real domain.
The functional $f(x)$ is the optimisation objective, and BO aims to minimise the objective score.
We use $x^*$ to denote the hyperparameters that can yield the lowest possible objective score, which is given by
\begin{equation}
  x^* = \arg\min_{x\in\mathcal{X}} f(x).
\end{equation}
Let $y=f(x)$ denote our performance measure in this study, where $y$ is typically \emph{F1 score} or \emph{accuracy} in our IoT cybersecurity detection domain.
We can then reformulate the BO optimisation objective to minimise the negative F1 score as our performance measure.

In a typical BO setting, we aim to find the conditional probability of the objective score given hyperparameters, i.e., $\mathbb{P}(y\,|\,x)$.
In TPE, we instead model $\mathbb{P}(x\,|\,y)$ and $\mathbb{P}(y)$ by transforming the generating process of hyperparameters and replacing the distribution of the hyperparameters prior with non-parametric densities.
TPE begins by collecting a few observations using a randomly selected set of hyperparameters.
Then, TPE sorts the collected observations by their respective objective score and divide them into groups based on quantile.
The quantiles are used to model empirical densities using Parzen Estimators.
A new sample of hyperparameters is then drawn from the densities, returning hyperparameters that yield the greatest expected improvements.
Therefore, TPE is an iterative process that uses the history of evaluated hyperparameters to create a probabilistic model, which is used to suggest the next set of hyperparameters to evaluate.

\subsection{Benchmarking Models}
This section presents the experimental details of each dataset's machine-learning models.
We evaluated 14 types of machine learning models, including six traditional ML models and eight ensemble models.
Each model has its respective set of hyperparameters.
We perform hyperparameters optimisation via a hierarchical Gaussian Process and a tree-structured Parzen estimator~\citep{bergstra2011_AlgoHype} with a budget of 45 trials to estimate the best performing set of parameters (see~\cref{sec:hyperparams} for visualisation of the influences of hyperparameters).

Our benchmarking models are as follows.
\begin{enumerate}
    \item \textit{Ridge Regressor (Ridge)}~\citep{hoerl1970_RidgRegr}: a model that addresses the classification problems as ordinary least squares by imposing a penalty on the size of the coefficients.
  Ridge treats the multi-class classification setup as a regression task with multiple outputs.
  It is a linear model that trains fast large dataset but cannot predicts well for nonlinear data.
\item \textit{Naive Bayes (NB)}~\citep{flach2004_NaivBaye}: is a model that uses Bayes theory for performing class conditional density estimation and with classes prior probability.
  The posterior class probability of input test data is derived using the Bayes theory to be assigned to the class with the maximum posterior class probability.
  NB can often learn quickly from a large dataset when compared to other classifiers; however, the conditional independence assumption in NB is rarely applicable to real-world problems.
\item \textit{Multi-layer Perceptron (MLP)}~\citep{gardner1998_ArtiNeur}: a feed-forward neural network that uses backpropagation for model training.
  MLP updates the weights between neurons to minimise the prediction error and can often learn nonlinear features and patterns within the dataset.
    MLP can often generalise well to new unseen data, but it is slow in convergence and often stuck in local minima.
\item \textit{Support Vector Machine (SVM)}~\citep{pisner2020_SuppVect}: a well-known model that classifies inputs by creating a hyperplane to separate each class.
    Due to its kernel trick of implicitly mapping the input features into higher-dimensional feature spaces, SVM can efficiently perform nonlinear classification problems.
\item \textit{Decision Tree (DT)}~\citep{myles2004_IntrDeci}: is a non-parametric model that uses a tree-like structure to construct its model for making decisions.
    A DT comprises a series of nodes and branches representing the decision rules inferred from the input features.
    The main benefit of DT lies in its interpretability due to its rule-based logic.
  \item \textit{K Nearest Neighbour (kNN)}~\citep{zhu2020_KNNbAppr}: is a classifier that simply stores the given input features and classifies input data via some similarity metrics.
    kNN, as a non-parametric method with a simple approach to the typical classification setup, is wildly adopted in many domains.

  \item \textit{Boostrap aggregation (Bagging)}~\citep{buhlmann2012_BaggBoos}: is an ensemble method that trains multiple weak classifiers on a random subset of the given dataset.
    Bagging can often reduce the variance within a noisy dataset, and the output of the Bagging model is obtained by averaging the predictions by its weak internal classifiers.
    In contrast to Boosting methods, the weaker classifiers in Bagging are trained independently, and Bagging can often avoid over-fitting in high-variance datasets.
  \item \textit{Adaptive Boosting (AdaBoost)}~\citep{hastie2009_MultAdab}: is a meta-classifier that works in conjunction with other learning algorithms.
    AdaBoost uses a weighted sum to combine the predicted output from other weak classifiers to output its final prediction.
    In particular, AdaBoost adaptively improves its performance under challenging classes by using additional weak learners to minimise the misclassification induced by previous classifiers.
  \item \textit{Random Forest (RF)}~\citep{pal2005_RandFore}: an ensemble method that internally combines the predictive power of multiple Decision Trees for outputting a final prediction.
    RF trains each DT using a random subset of the input features and uses bootstrapping approach by subsampling random features with replacement.
  \item \textit{Extremely Randomised Trees (ERT)}~\citep{geurts2006_ExtrRand}: an ensemble method that behaves similarly to RF, which trains multiple weak decision trees as the weak learner.
    However, unlike RF, each decision tree is trained with the whole dataset instead of subsampling.
    ERT also randomly select the split point to split nodes in decision trees instead of finding the optimal split as in RF.
    ERT can often train faster than RF and attains a lower overall variance due to the random splitting of nodes.
  \item \textit{Gradient Boosting Machine (GBM)}~\citep{friedman2001_GreeFunc}: is an ensemble technique that uses multiple weak prediction models built in a stage-wise fashion.
    Similar to RF, a decision tree is a common choice for being the weak prediction model within GB.
    However, in GB, each weak predictor is trained to correct the residuals of its predecessor, and as a result, it can often achieve a lower model bias than RF.
  \item \textit{Extreme Gradient Boosting (XGB)}~\citep{chen2016_XgboScal}: a tree boosting method that uses the same idea of gradient boosting.
    In contrast to GB, XGB uses the second-order derivatives method to find the optimal constant in each terminal node and uses regularisation of the tree to avoid overfitting.
\item \textit{Voting}~\citep{ruta2005_ClasSele}: an ensemble method that uses voting to combine the predicted outputs from several individual predictors.
  The Voting ensemble can use the majority vote or the average predicted probability to predict the class labels.
    The Voting ensemble can incorporate arbitrary predictors and often balance out the weakness between the nested predictors.
  \item \textit{Stacked Generalisation (Stacking)}~\citep{naimi2018_StacGene}: an ensemble method that allows one to combine several different prediction algorithms by stacking the individual predictors' output and uses a final classifier to compute a final prediction.
  The stacking approach is straightforward, and different kinds of predictors can be easily combined, which could potentially use the strength of some models to compensate for the weaknesses of other models.
\end{enumerate}

\section{Experiments}

In this section, we present the experimental results from benchmarking the discussed Machine Learning models against various IoT anomaly detection datasets.

\subsection{Experimental Setups}

We randomly split the dataset for each target IoT anomaly detection dataset by using $20\%$ of the data as the unseen test set.
Hyperparameter optimisation is then performed by using the training set for testing various hyperparameters, thereby computing the corresponding F1-score with 5-fold cross-validation.
During the hyperparameter optimisation process, we kept the test set hidden along with the 5-fold cross-validation to ensure that the optimised hyperparameters can perform well robustly across a wide range of domain subsets.
This procedure is repeated 45 times, and we use the best-performing hyperparameters under this cross-validation to report the following scores on the test set.

The reported score is computed using the predictive results of the models using the tunned hyperparameters.
The experimental results are performed on an HPC cluster with 32 requested CPU cores of Intel Xeon E5-2680 V3 2.50GHz processor, 128GB RAM, and NVIDIA V100 SXM2 GPU equipped with 16GB vRAM.
In total, this study performed 5-fold cross-validation $\times$ 45 trials $\times$ 14 models $\times$ 5 datasets, with a total of 15750 model training episodes.

\begin{table*}[!bt]
  \centering
  \caption{\exprresultscaption{} \label{table:expr-results}}
  \begin{tabular}{rcllllll}
    \toprule
    \multicolumn{2}{c}{\multirow{2}{*}{Model~~~~~Ensemble?}} & \multicolumn{1}{c}{\multirow{2}{*}{Dataset}} & \multicolumn{1}{c}{\multirow{2}{*}{Type}} & \multicolumn{4}{c}{Metric}                                                                                         \\ \cmidrule(l){5-8}
    \multicolumn{2}{c}{}                                     & \multicolumn{1}{c}{}                         & \multicolumn{1}{c}{}                      & \multicolumn{1}{c}{Accuracy} & \multicolumn{1}{c}{Precision} & \multicolumn{1}{c}{Recall} & \multicolumn{1}{c}{F1} \\ \midrule
\multirow{5}{*}{Ridge}     & \multirow{5}{*}{}           & \multirow{3}{*}{IoTID20}                     & \texttt{Binary}                           & 0.94                         & 0.83                          & 0.55                       & 0.57                   \\
                           &                             &                                              & Multi-Cat                                 & 0.77                         & 0.78                          & 0.50                       & 0.54                   \\
                           &                             &                                              & Multi-SubCat                              & 0.54                         & 0.43                          & 0.35                       & 0.32                   \\
                           &                             & \multirow{2}{*}{IoT23}                       & Binary                                    & 0.86                         & 0.43                          & 0.50                       & 0.46                   \\
                           &                             &                                              & Multi-Cat                                 & 0.81                         & 0.78                          & 0.54                       & 0.57                   \\ \midrule
\multirow{5}{*}{NB}        & \multirow{5}{*}{}           & \multirow{3}{*}{IoTID20}                     & Binary                                    & 0.45                         & 0.55                          & 0.69                       & 0.39                   \\
                           &                             &                                              & Multi-Cat                                 & 0.57                         & 0.47                          & 0.58                       & 0.47                   \\
                           &                             &                                              & Multi-SubCat                              & 0.46                         & 0.34                          & 0.36                       & 0.32                   \\
                           &                             & \multirow{2}{*}{IoT23}                       & Binary                                    & 0.81                         & 0.45                          & 0.48                       & 0.46                   \\
                           &                             &                                              & Multi-Cat                                 & 0.43                         & 0.32                          & 0.35                       & 0.31                   \\ \midrule
\multirow{5}{*}{MLP}       & \multirow{5}{*}{}           & \multirow{3}{*}{IoTID20}                     & Binary                                    & 0.95                         & 0.95                          & 0.60                       & 0.65                   \\
                           &                             &                                              & Multi-Cat                                 & 0.78                         & 0.84                          & 0.53                       & 0.57                   \\
                           &                             &                                              & Multi-SubCat                              & 0.50                         & 0.45                          & 0.39                       & 0.36                   \\
                           &                             & \multirow{2}{*}{IoT23}                       & Binary                                    & 0.86                         & 0.43                          & 0.50                       & 0.46                   \\
                           &                             &                                              & Multi-Cat                                 & 0.57                         & 0.06                          & 0.11                       & 0.08                   \\ \midrule
\multirow{5}{*}{SVM}       & \multirow{5}{*}{}           & \multirow{3}{*}{IoTID20}                     & Binary                                    & 0.96                         & 0.91                          & 0.71                       & 0.78                   \\
                           &                             &                                              & Multi-Cat                                 & 0.80                         & 0.80                          & 0.57                       & 0.62                   \\
                           &                             &                                              & Multi-SubCat                              & 0.58                         & 0.49                          & 0.42                       & 0.38                   \\
                           &                             & \multirow{2}{*}{IoT23}                       & Binary                                    & 0.90                         & 0.95                          & 0.62                       & 0.67                   \\
                           &                             &                                              & Multi-Cat                                 & 0.49                         & 0.42                          & 0.41                       & 0.43                   \\ \midrule
\multirow{5}{*}{DT}        & \multirow{5}{*}{}           & \multirow{3}{*}{IoTID20}                     & Binary                                    & 0.96                         & 0.91                          & 0.73                       & 0.79                   \\
                           &                             &                                              & Multi-Cat                                 & 0.76                         & 0.31                          & 0.40                       & 0.35                   \\
                           &                             &                                              & Multi-SubCat                              & 0.58                         & 0.39                          & 0.42                       & 0.39                   \\
                           &                             & \multirow{2}{*}{IoT23}                       & Binary                                    & 0.86                         & 0.43                          & 0.50                       & 0.46                   \\
                           &                             &                                              & Multi-Cat                                 & 0.64                         & 0.18                          & 0.20                       & 0.18                   \\ \midrule
\multirow{5}{*}{kNN}       & \multirow{5}{*}{}           & \multirow{3}{*}{IoTID20}                     & Binary                                    & 0.98                         & 0.96                          & 0.90                       & 0.93                   \\
                           &                             &                                              & Multi-Cat                                 & 0.85                         & 0.81                          & 0.77                       & 0.78                   \\
                           &                             &                                              & Multi-SubCat                              & 0.60                         & 0.56                          & 0.51                       & 0.52                   \\
                           &                             & \multirow{2}{*}{IoT23}                       & Binary                                    & 0.92                         & 0.89                          & 0.85                       & 0.87                   \\
                           &                             &                                              & Multi-Cat                                 & 0.71                         & 0.23                          & 0.15                       & 0.20                   \\ \midrule
\multirow{5}{*}{Bagging}   & \multirow{5}{*}{\checkmark} & \multirow{3}{*}{IoTID20}                     & Binary                                    & 0.99                         & 0.99                          & 0.93                       & 0.96                   \\
                           &                             &                                              & Multi-Cat                                 & 0.87                         & 0.84                          & 0.80                       & 0.81                   \\
                           &                             &                                              & Multi-SubCat                              & 0.65                         & 0.65                          & 0.55                       & 0.56                   \\
                           &                             & \multirow{2}{*}{IoT23}                       & Binary                                    & 1.00                         & 1.00                          & 1.00                       & 1.00                   \\
                           &                             &                                              & Multi-Cat                                 & 0.85                         & 0.51                          & 0.43                       & 0.49                   \\ \midrule
\multirow{5}{*}{AdaBoost}  & \multirow{5}{*}{\checkmark} & \multirow{3}{*}{IoTID20}                     & Binary                                    & 0.99                         & 0.98                          & 0.89                       & 0.93                   \\
                           &                             &                                              & Multi-Cat                                 & 0.81                         & 0.85                          & 0.63                       & 0.66                   \\
                           &                             &                                              & Multi-SubCat                              & 0.61                         & 0.60                          & 0.46                       & 0.46                   \\
                           &                             & \multirow{2}{*}{IoT23}                       & Binary                                    & 0.99                         & 0.99                          & 0.98                       & 0.99                   \\
                           &                             &                                              & Multi-Cat                                 & 0.84                         & 0.36                          & 0.32                       & 0.33                   \\ \midrule
    \end{tabular}
\end{table*}

\begin{table*}[!bt]
  \ContinuedFloat
  \centering
  \caption{\emph{Continued:} \exprresultscaption{} \label{table:expr-results-cont}}
  \begin{tabular}{@{}rcllllll@{}}
    \toprule
    \multicolumn{2}{c}{\multirow{2}{*}{Model~~~~~Ensemble?}} & \multicolumn{1}{c}{\multirow{2}{*}{Dataset}} & \multicolumn{1}{c}{\multirow{2}{*}{Type}} & \multicolumn{4}{c}{Metric}                                                                                         \\ \cmidrule(l){5-8}
    \multicolumn{2}{c}{}                                     & \multicolumn{1}{c}{}                         & \multicolumn{1}{c}{}                      & \multicolumn{1}{c}{Accuracy} & \multicolumn{1}{c}{Precision} & \multicolumn{1}{c}{Recall} & \multicolumn{1}{c}{F1} \\ \midrule
\multirow{5}{*}{RF}        & \multirow{5}{*}{\checkmark} & \multirow{3}{*}{IoTID20} & Binary                                    & 0.94                         & 0.47                          & 0.50                       & 0.48                   \\
                           &                             &                          & Multi-Cat                                 & 0.66                         & 0.13                          & 0.20                       & 0.16                   \\
                           &                             &                          & Multi-SubCat                              & 0.29                         & 0.03                          & 0.11                       & 0.05                   \\
                           &                             & \multirow{2}{*}{IoT23}   & Binary                                    & 0.86                         & 0.43                          & 0.50                       & 0.46                   \\
                           &                             &                          & Multi-Cat                                 & 0.57                         & 0.06                          & 0.11                       & 0.08                   \\ \midrule
\multirow{5}{*}{ERT}       & \multirow{5}{*}{\checkmark} & \multirow{3}{*}{IoTID20} & Binary                                    & 0.94                         & 0.47                          & 0.50                       & 0.48                   \\
                           &                             &                          & Multi-Cat                                 & 0.75                         & 0.35                          & 0.39                       & 0.37                   \\
                           &                             &                          & Multi-SubCat                              & 0.52                         & 0.22                          & 0.31                       & 0.25                   \\
                           &                             & \multirow{2}{*}{IoT23}   & Binary                                    & 0.97                         & 0.98                          & 0.89                       & 0.93                   \\
                           &                             &                          & Multi-Cat                                 & 0.84                         & 0.30                          & 0.31                       & 0.30                   \\ \midrule
\multirow{5}{*}{GBM}       & \multirow{5}{*}{\checkmark} & \multirow{3}{*}{IoTID20} & Binary                                    & 0.95                         & 0.96                          & 0.63                       & 0.69                   \\
                           &                             &                          & Multi-Cat                                 & 0.83                         & 0.85                          & 0.69                       & 0.70                   \\
                           &                             &                          & Multi-SubCat                              & 0.61                         & 0.53                          & 0.48                       & 0.48                   \\
                           &                             & \multirow{2}{*}{IoT23}   & Binary                                    & 0.86                         & 0.43                          & 0.50                       & 0.46                   \\
                           &                             &                          & Multi-Cat                                 & 0.89                         & 0.38                          & 0.38                       & 0.38                   \\ \midrule
\multirow{5}{*}{XGB}       & \multirow{5}{*}{\checkmark} & \multirow{3}{*}{IoTID20} & Binary                                    & 0.99                         & 0.99                          & 0.94                       & 0.96                   \\
                           &                             &                          & Cat                                       & 0.87                         & 0.84                          & 0.81                       & 0.81                   \\
                           &                             &                          & SubCat                                    & 0.66                         & 0.65                          & 0.56                       & 0.57                   \\
                           &                             & \multirow{2}{*}{IoT23}   & Binary                                    & 1.00                         & 1.00                          & 1.00                       & 1.00                   \\
                           &                             &                          & Multi                                     & 0.98                         & 0.74                          & 0.63                       & 0.66                   \\ \midrule
\multirow{5}{*}{Stacking}  & \multirow{5}{*}{\checkmark} & \multirow{3}{*}{IoTID20} & Binary                                    & 0.99                         & 0.98                          & 0.94                       & 0.96                   \\
                           &                             &                          & Multi-Cat                                 & 0.87                         & 0.83                          & 0.78                       & 0.80                   \\
                           &                             &                          & Multi-SubCat                              & 0.67                         & 0.65                          & 0.57                       & 0.58                   \\
                           &                             & \multirow{2}{*}{IoT23}   & Binary                                    & 0.88                         & 0.85                          & 0.90                       & 0.87                   \\
                           &                             &                          & Multi-Cat                                 & 0.59                         & 0.32                          & 0.36                       & 0.35                   \\ \midrule
\multirow{5}{*}{Voting}    & \multirow{5}{*}{\checkmark} & \multirow{3}{*}{IoTID20} & Binary                                    & 0.98                         & 0.95                          & 0.93                       & 0.94                   \\
                           &                             &                          & Multi-Cat                                 & 0.85                         & 0.84                          & 0.74                       & 0.77                   \\
                           &                             &                          & Multi-SubCat                              & 0.63                         & 0.61                          & 0.54                       & 0.53                   \\
                           &                             & \multirow{2}{*}{IoT23}   & Binary                                    & 0.99                         & 0.99                          & 0.95                       & 0.97                   \\
                           &                             &                          & Multi-Cat                                 & 0.82                         & 0.39                          & 0.41                       & 0.41                   \\ \bottomrule
    \end{tabular}
\end{table*}

\subsection{Evaluation Metrics}
In this study, we use \emph{accuracy}, \emph{precision}, \emph{recall}, and \emph{F1-score} to evaluate our models.
In the classification problem, there are four possible results in the model prediction: true positive (TP), true negative (TN), false positive (FP), and false negative (FN).
TP refers to the number of instances that have been correctly identified as normal.
TN represents the number of instances that are classified correctly as malicious.
FP represents the number of malicious instances that are wrongly classified as normal.
FN refers to the number of instances that misclassify normal data as malicious data.
Our evaluation metrics are then given by
\begin{align}
  \text{Accuracy} &= \frac{\text{TP} + \text{TN}}{\text{TP} + \text{TN} + \text{FP} + \text{FN}} \\
  \text{Precision} &= \frac{\text{TP}}{\text{TP} + \text{FP}} \\
  \text{Recall} &= \frac{\text{TP}}{\text{TP} + \text{FN}} \\
  \text{F1-score} &= \frac{2 \times (\text{Precision} \times \text{Recall})}{\text{Precision} + \text{Recall}}
  .
\end{align}
Accuracy is an intuitive measure summarising the ratio of correctly predicted observation to total observation.
However, when the dataset contains an asymmetric number of classes, which is the case for some of our datasets, accuracy cannot convey the whole picture of model performance.
Precision refers to the ratio of correctly predicted positive observations to the total predicted observations.
A high value in precision denotes a low false positive rate.
Recall, sometimes referred to as sensitivity, is the ratio of correctly predicted positive observation to the total number of that class.
Recall answers the percentage of correct positives our model can capture.
The F1 score is a weighted average of precision and recall.
F1 score is usually more helpful information than accuracy, especially in datasets with an uneven class distribution.

\subsection{Experimental Results}\label{sec:model-results}

All models are trained on the same training set and evaluated on the same test set, and we use five-fold cross-validation on the train set to select the set of hyperparameters that achieved the highest performance in the F1 score.
The numerical results from the models with the best-performing hyperparameters are summarised and presented in~\cref{table:expr-results}.

The results are indicative of the overall performance of the various approaches.
Traditional linear models, such as Ridge, perform poorly across all five datasets.
The reason is likely due to the limited expressiveness of the model to learn the nonlinearity input features.
Models such as DT can achieve a decent result on more straightforward datasets with binary labels, but their performance drops rapidly once the dataset contains more label classes.
Moreover, models like RF exhibit a higher accuracy value but with a considerably low precision value, indicating that the model is biased towards the classes with a higher number of instances leading to a high false positive rate.

In contrast, ensemble models can utilise multiple weaker predictors' predictive power to improve the overall predictive accuracy.
For example, XGB can achieve superior results on most datasets due to its utilisation of multiple DT under the hood.
Most ensemble models can consistently achieve higher model performance across a wide range of IoT anomaly scenarios.
For example, ensemble models like Stacking and XGB have consistently high performance across the more trivial dataset \texttt{IoTID20 Binary} and the more complicated \texttt{IoTID20 Multi-SubCat}.
When the number of classes increases, most traditional models like RF have a dramatic drop in performance, while ensemble models are still able to maintain performance in the higher end of the spectrum

\subsection{Hyperparameters Optimisation}\label{sec:hyperparams}

Hyperparameters tend to be one of the aspects that can mystify ML due to models' sensitiveness towards the choice of hyperparameters.
A decent model can achieve mediocre results due to a poor choice of hyperparameters.
In this section, we study the models' sensitivity toward their choice of hyperparameters.
All of the results are obtained via 45 trials of hyperparameter optimisation for each of the models.
As mentioned earlier, the tree-structured Parzen estimator~\citep{bergstra2011_AlgoHype} is used for our Bayesian Optimisation procedure by fitting multiple Gaussian Mixture Models for searching the most promising hyperparameter values.
In the following, we focus on the multi-class anomaly detection dataset \texttt{IoTID20 Multi-SubCat} and provide visualisation of the hyperparameter optimisation procedure.

\begin{figure*}[!tb]
  \begin{subfigure}[t]{.475\textwidth}
    \includegraphics[width=\linewidth, trim={0 0 4.3cm 0.5cm}, clip]{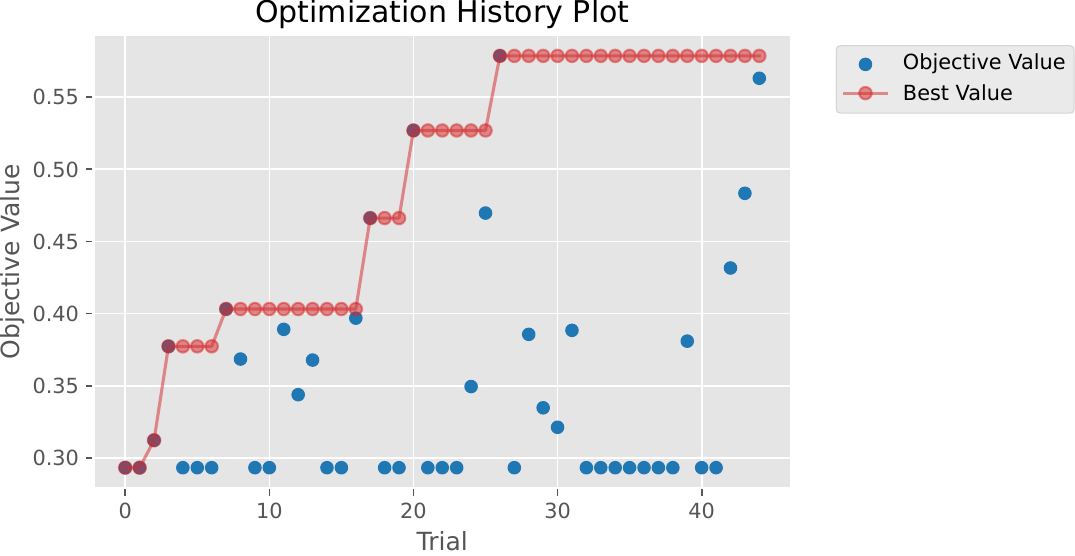}
    \caption{\centering \scriptsize Decision Tree (DT)}
  \end{subfigure}%
  \begin{subfigure}[t]{.475\textwidth}
    \includegraphics[width=\linewidth, trim={0 0 4.3cm 0.5cm}, clip]{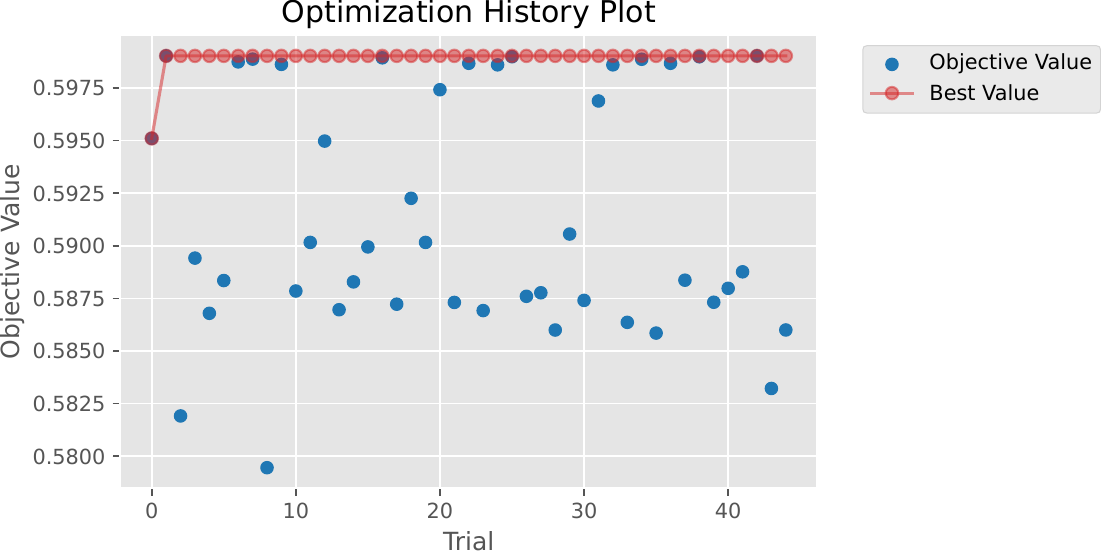}
    \caption{\centering \scriptsize K Nearest Neighbour (kNN)}
  \end{subfigure}\vspace{.7em}
  \begin{subfigure}[t]{.475\textwidth}
    \includegraphics[width=\linewidth, trim={0 0 4.3cm 0.5cm}, clip]{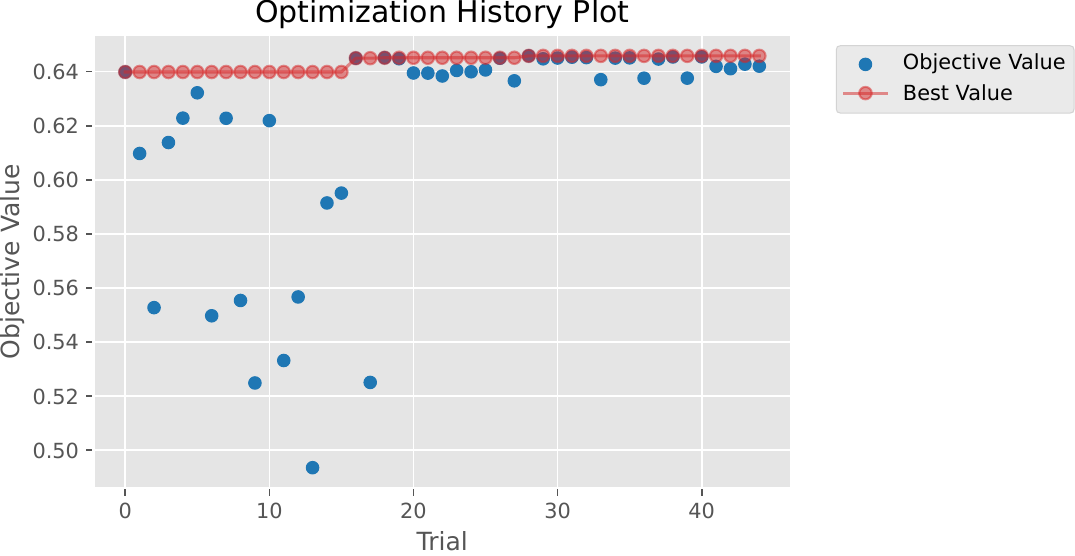}
    \caption{\centering \scriptsize Boostrap aggregation (Bagging)}
  \end{subfigure}%
  \begin{subfigure}[t]{.475\textwidth}
    \includegraphics[width=\linewidth, trim={0 0 4.3cm 0.5cm}, clip]{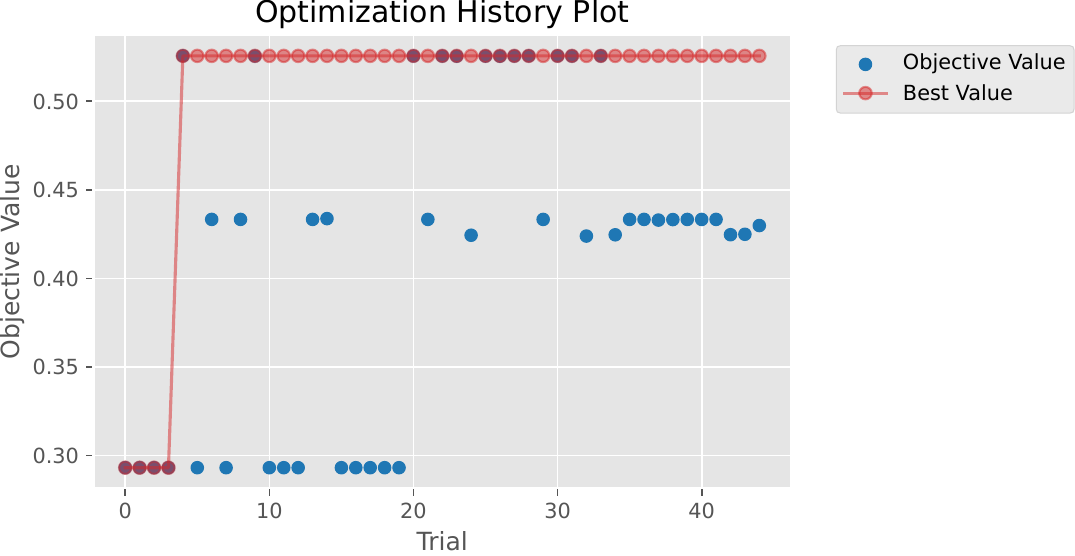}
    \caption{\centering \scriptsize Extremely Randomised Trees (ERT)}
  \end{subfigure}\vspace{.7em}
  \begin{subfigure}[t]{.475\textwidth}
    \includegraphics[width=\linewidth, trim={0 0 4.3cm 0.5cm}, clip]{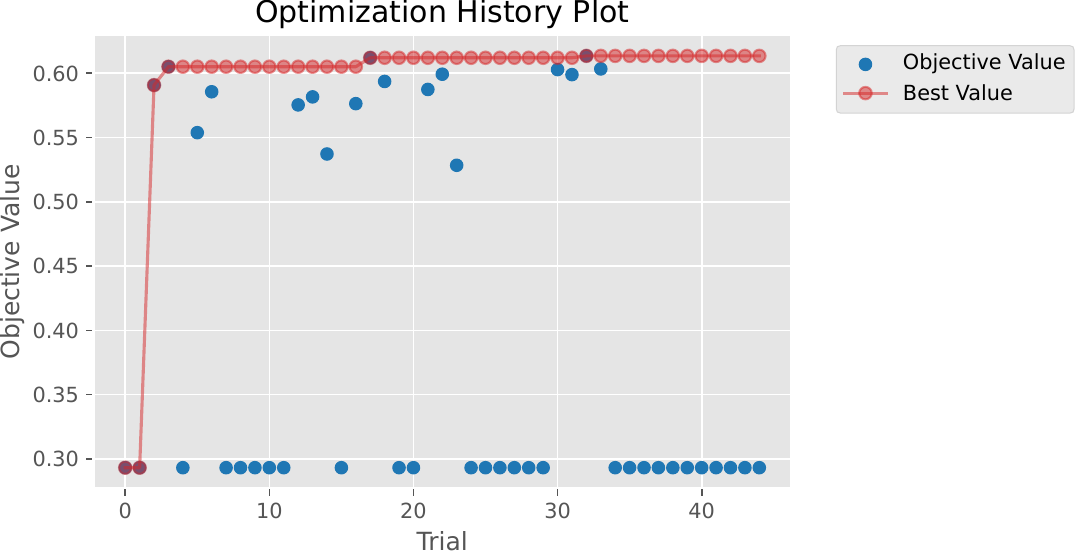}
    \caption{\centering \scriptsize Gradient Boosting Machine (GBM)}
  \end{subfigure}%
  \begin{subfigure}[t]{.475\textwidth}
    \includegraphics[width=\linewidth, trim={0 0 4.3cm 0.5cm}, clip]{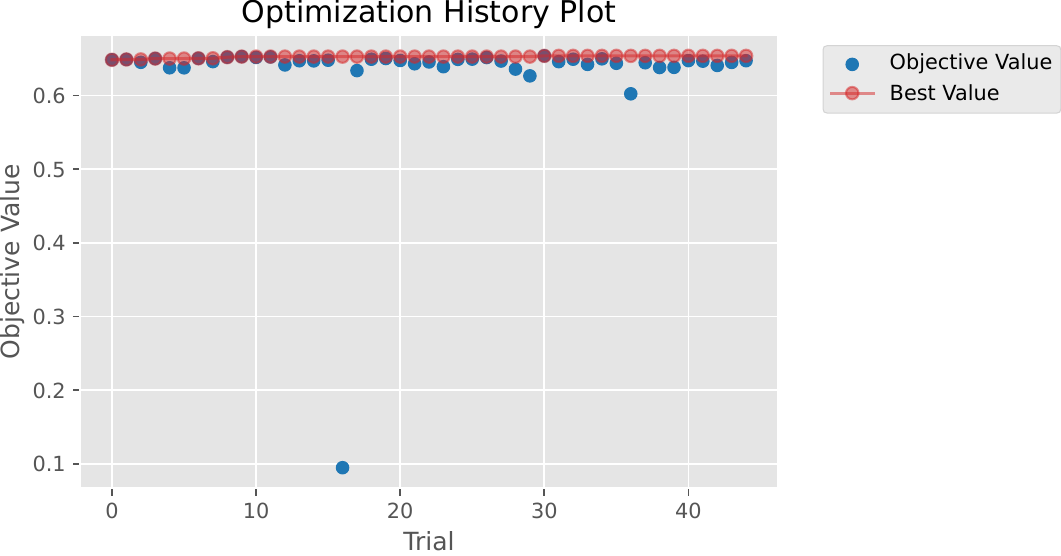}
    \caption{\centering \scriptsize Extreme Gradient Boosting (XGB)}
  \end{subfigure}\vspace{.7em}
  \caption{Plots of the history of hyperparameter optimisation process against the achieved F1 objective values.
  \label{fig:optimisation-history}}
\end{figure*}

\subsubsection{Visualising Optimisation History}

\begin{figure*}[!tbh]
  \centering
  \begin{subfigure}[t]{.475\textwidth}
    \includegraphics[width=\linewidth, trim={0 0 0 0.6cm}, clip]{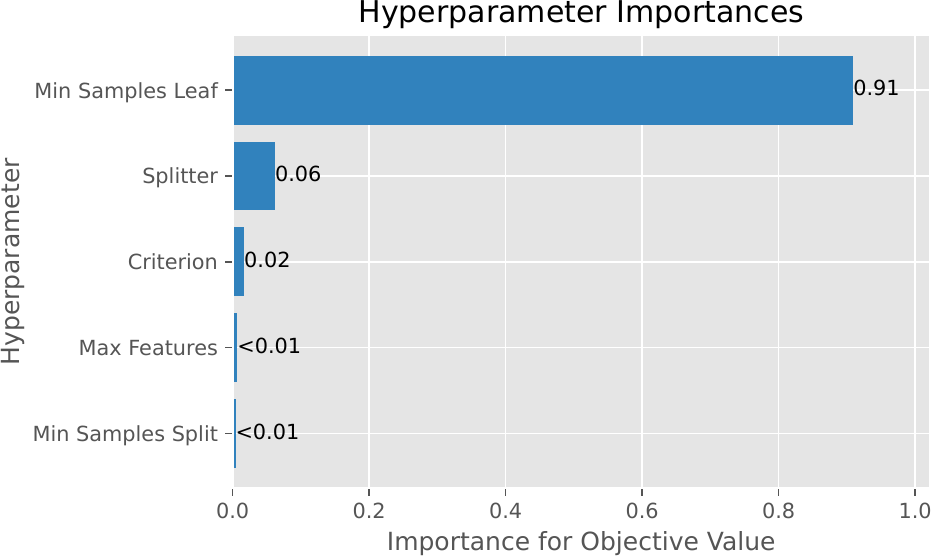}
    \caption{\centering \scriptsize Decision Tree (DT)}
  \end{subfigure}%
  \begin{subfigure}[t]{.475\textwidth}
    \includegraphics[width=\linewidth, trim={0 0 0 0.6cm}, clip]{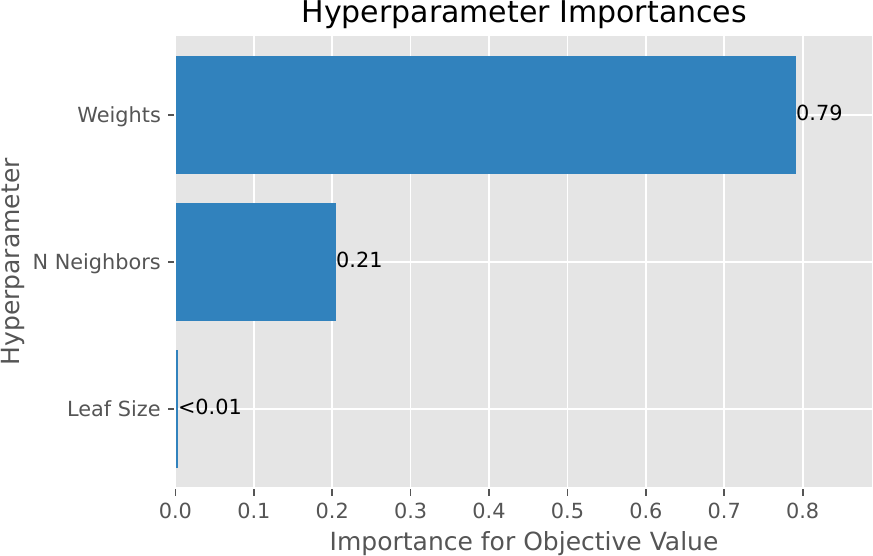}
    \caption{\centering \scriptsize K Nearest Neighbour (kNN)}
  \end{subfigure}\vspace{.7em}
  \begin{subfigure}[t]{.475\textwidth}
    \includegraphics[width=\linewidth, trim={0 0 0 0.6cm}, clip]{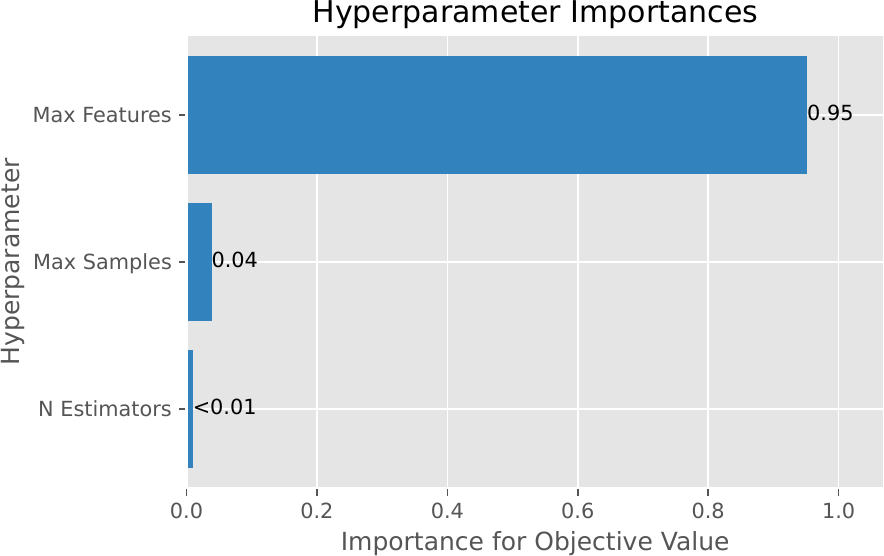}
    \caption{\centering \scriptsize Boostrap aggregation (Bagging)}
  \end{subfigure}%
  \begin{subfigure}[t]{.475\textwidth}
    \includegraphics[width=\linewidth, trim={0 0 0 0.6cm}, clip]{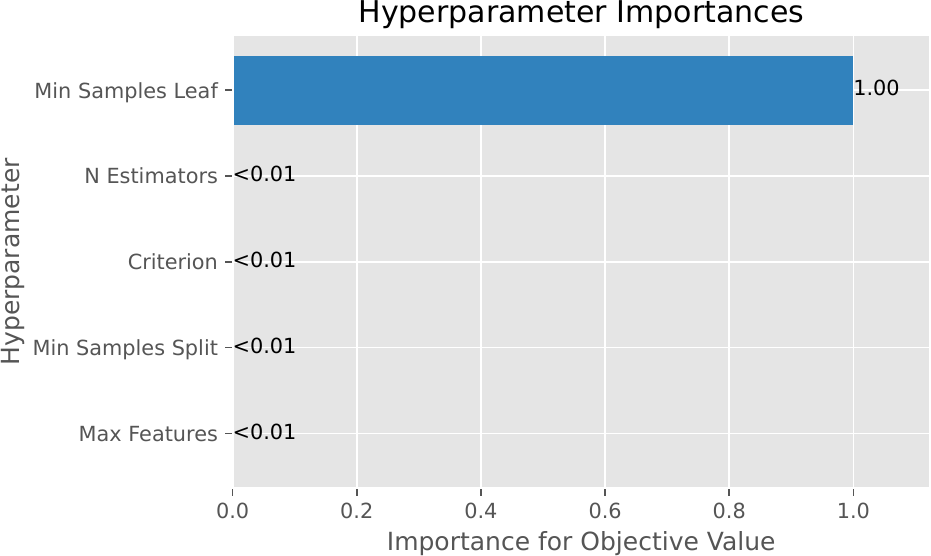}
    \caption{\centering \scriptsize Extremely Randomised Trees (ERT)}
  \end{subfigure}\vspace{.7em}
  \begin{subfigure}[t]{.475\textwidth}
    \includegraphics[width=\linewidth, trim={0 0 0 0.6cm}, clip]{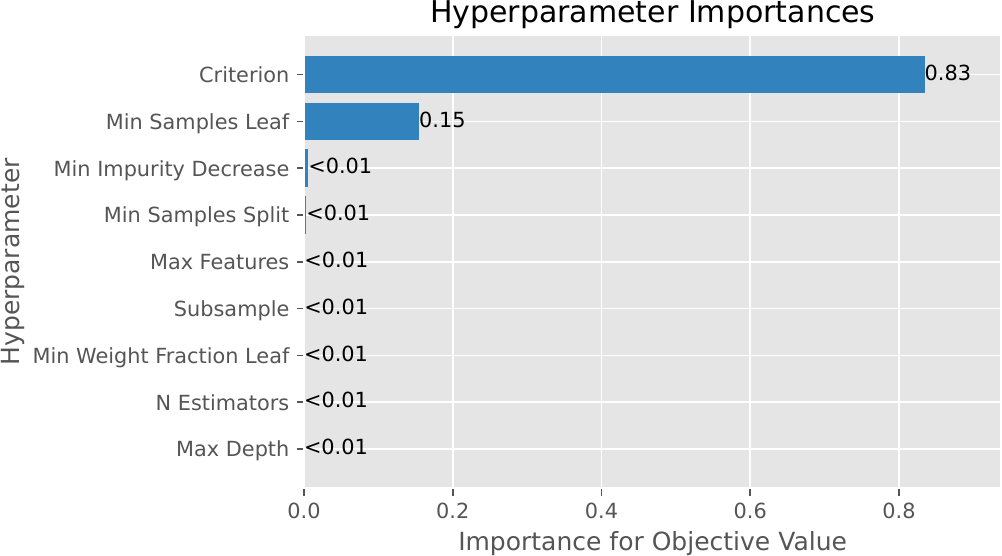}
    \caption{\centering \scriptsize Gradient Boosting Machine (GBM)}
  \end{subfigure}%
  \begin{subfigure}[t]{.475\textwidth}
    \includegraphics[width=\linewidth, trim={0 0 0 0.6cm}, clip]{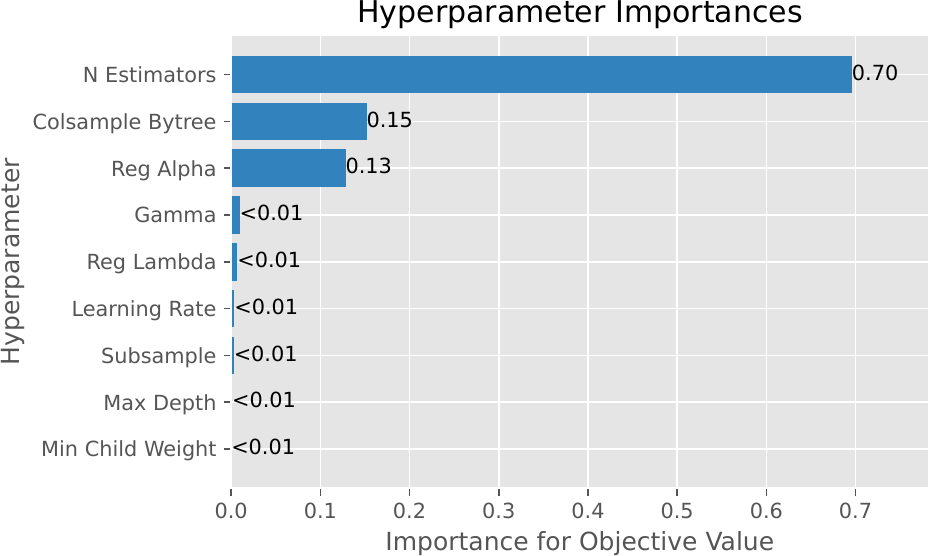}
    \caption{\centering \scriptsize Extreme Gradient Boosting (XGB)}
  \end{subfigure}\vspace{.7em}
  \caption{Plots of quantifying the importance of each hyperparameter with respect to the objective value via the functional ANOVA framework.
  \label{fig:param-importance}}
\end{figure*}

The hyperparameter optimisation history of a few selected models is shown in~\cref{fig:optimisation-history}, which includes a few traditional ML models and ensemble models.
The figure illustrates the optimisation history of the object value, which is the F1 score obtained with a 5-fold cross-validation on the training set.
The plots include the history of the 45 trials horizon and keep track of the best-obtained value.
The scattering of the objective value also illustrates the consistency of each method.
For example, traditional methods such as DT can achieve an F1 score above 0.55 in their training set.
However, its optimisation history has a significant variance.
Moreover, the same model only obtains an F1 score of 0.39 in its test set (see~\cref{table:expr-results-cont}).
The result suggests that models like DT are prone to overfitting on their training set despite using k-fold cross-validation.

On the other hand, ensemble models seem to have less variance in their optimisation history.
The extreme gradient boosting model contains an outlier that achieves poor performance (around 0.1); however, generally, most attainable objective values tend to fluctuate in the higher end of the spectrum.
As a result, ensemble models are less prone to the choice of hyperparameters.
Poor choices of hyperparameters can still lead to inferior performance.
However, the end results from ensemble models are still more consistent than traditional results due to the way that ensemble models aggregate results from their internal estimators.

\begin{figure*}[!htb]
  \centering
  \begin{subfigure}[t]{.95\textwidth}
    \hfill \includegraphics[width=\linewidth, trim={0 0 0 1cm}, clip]{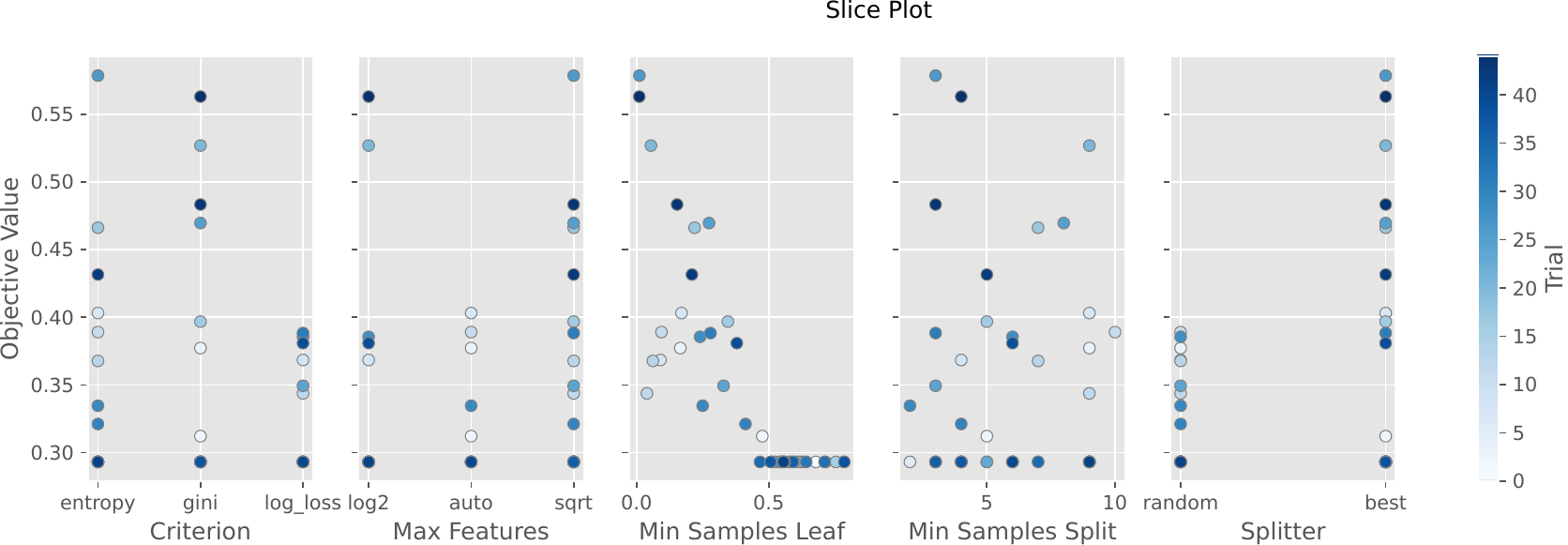}
    \caption{\centering \scriptsize Decision Tree (DT)}
  \end{subfigure}\vspace{1em}
  \begin{subfigure}[t]{.95\textwidth}
    \hfill \includegraphics[width=.7\linewidth, trim={0 0 0 1cm}, clip]{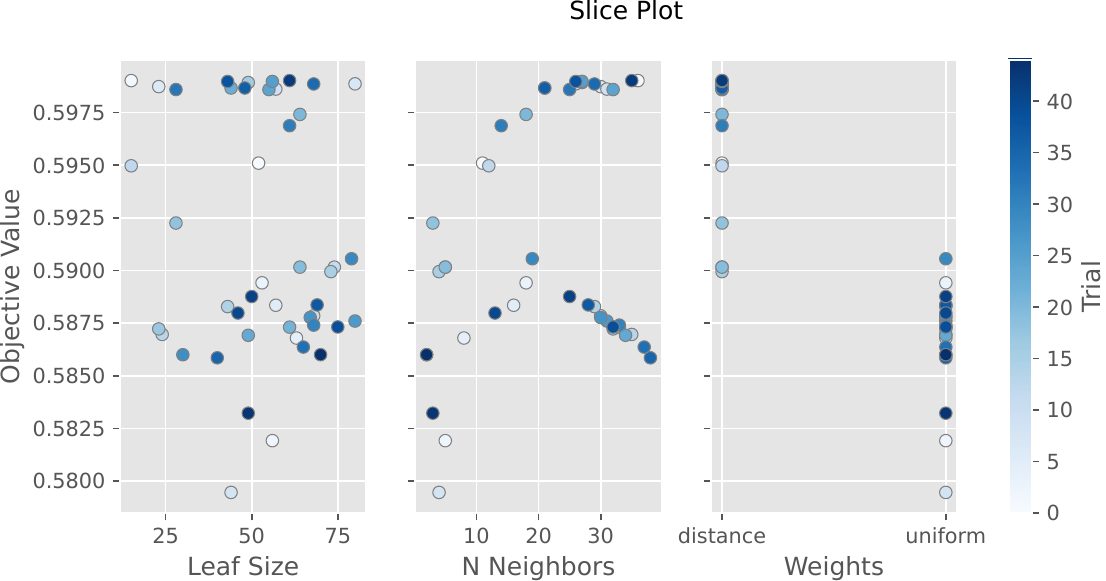}
    \caption{\centering \scriptsize K Nearest Neighbour (kNN)}
  \end{subfigure}\vspace{1em}
  \begin{subfigure}[t]{.95\textwidth}
    \hfill \includegraphics[width=.7\linewidth, trim={0 0 0 1cm}, clip]{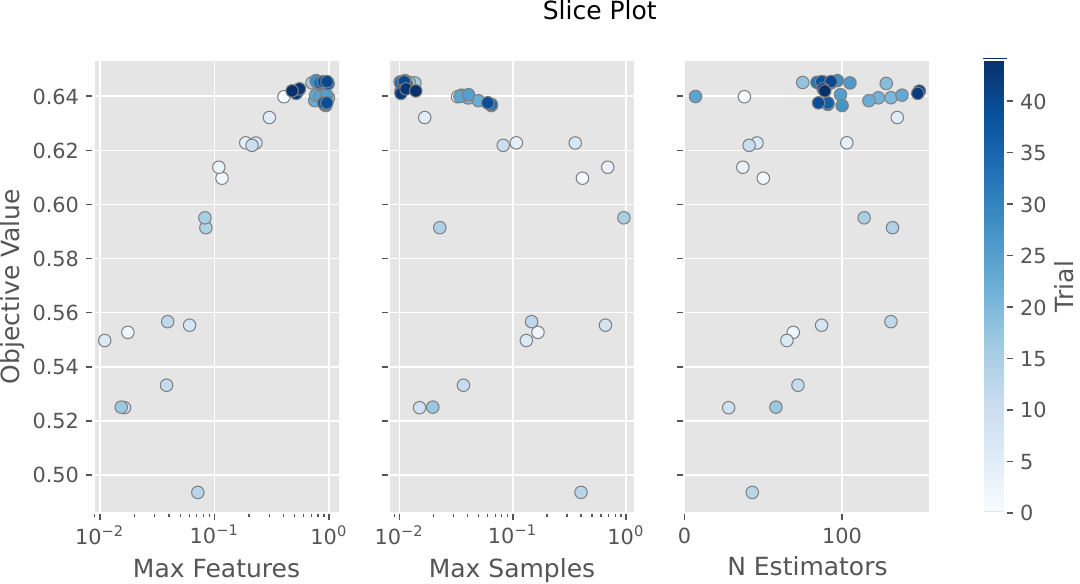}
    \caption{\centering \scriptsize Boostrap aggregation (Bagging)\label{fig:parameter-relationship:bagging}}
  \end{subfigure}\vspace{1em}
  \caption{Slice plot of parameter relationship with respect to the F1 objective value.
  \label{fig:parameter-relationship}}
\end{figure*}

\begin{figure*}[!htb]
  \ContinuedFloat
  \centering
  \begin{subfigure}[t]{.95\textwidth}
    \hfill \includegraphics[width=.9\linewidth, trim={0 0 0 1cm}, clip]{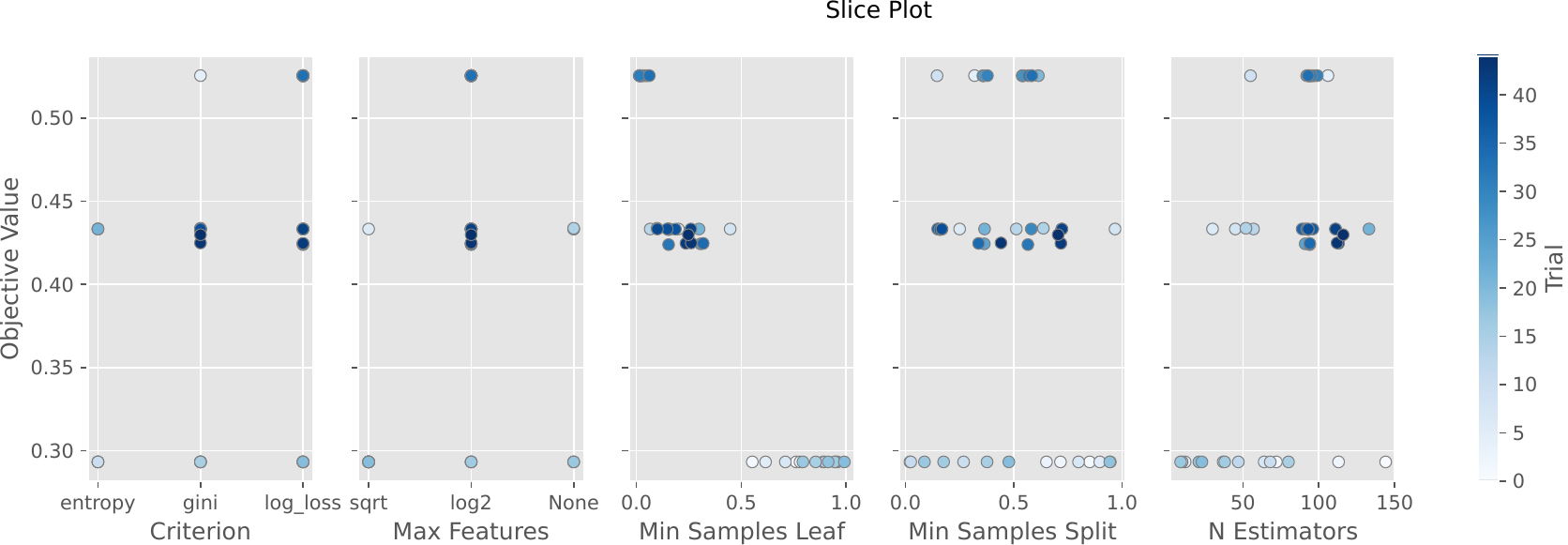}
    \caption{\centering \scriptsize Extremely Randomised Trees (ERT)}
  \end{subfigure}\vspace{1em}
  \begin{subfigure}[t]{.95\textwidth}
    \hfill \includegraphics[width=\linewidth, trim={0 0 4.1cm 1cm}, clip]{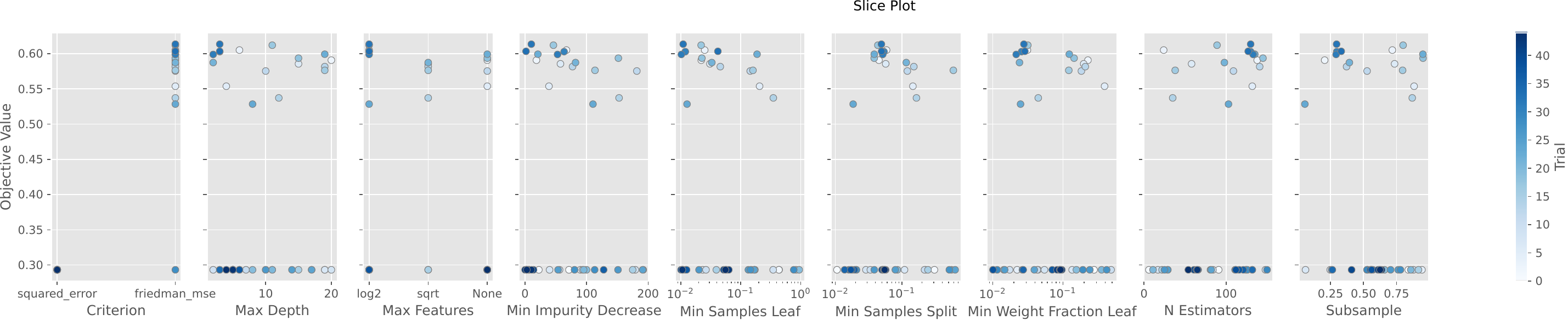}
    \caption{\centering \scriptsize Gradient Boosting Machine (GBM)\label{fig:parameter-relationship:gbm}}
  \end{subfigure}\vspace{1em}
  \begin{subfigure}[t]{.95\textwidth}
    \hfill \includegraphics[width=\linewidth, trim={0 0 4.1cm 1cm}, clip]{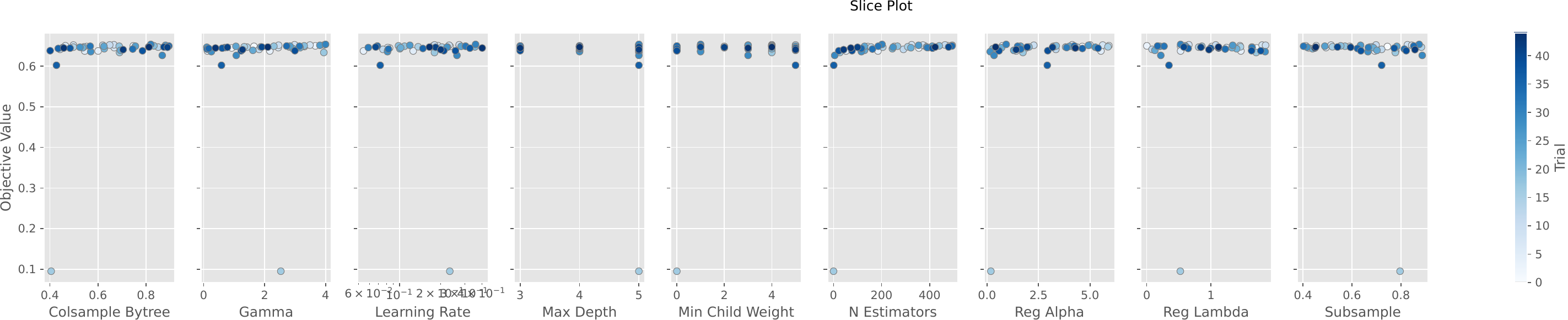}
    \caption{\centering \scriptsize Extreme Gradient Boosting (XGB)}
  \end{subfigure}\vspace{1em}
  \caption{\emph{Continued:} Slice plot of parameter relationship with respect to the F1 objective value.
  \label{fig:parameter-relationship-cont}}
\end{figure*}

\subsubsection{Estimating the Importance of Hyperparameter}

The performance of the evaluated methods can be heavily dependent on the hyperparameter settings.
Therefore, in our parametric study, we leverage a functional ANOVA framework to quantify the importance of the hyperparameters used in the various modes~\citep{hutter2014_EffiAppr}.
The estimation leverages random forest models to analyse the target model's variance, which helps decompose the importance of its corresponding hyperparameters.

\Cref{fig:parameter-relationship} illustrates the critical features in our best-performing model.
This measure is obtained by updating the hyperparameters and evaluating the model F1 score whilst treating the model as a black-box function.
The fANOVA framework~\citep{hutter2014_EffiAppr} is then used to estimate how the changes of each hyperparameter affect the final F1 score on the training set.
The degree of importance can have a high variation among hyperparameters, and the numerical value presented in~\cref{fig:parameter-relationship} can be seen as the proportion of effect that the hyperparameter exhibit on the final F1 score.

We can observe common patterns among various models.
For example, the ``Min Samples Leaf'' hyperparameter in both tree-based methods (DT and ERT) is the most contributing factor.
The parameter controls the minimum number of samples required to be a leaf node, which determines the resulting structure of the trees.
Similarly, we can see that parameters that control the number of features or estimators also contribute tremendously to the F1 score (``Max Features'' in Bagging and ``N Estimators'' in XGB).
The former is the maximum threshold of features used to train its internal sub-estimator, while the latter directly controls the number of sub-estimators.
Finally, the ``Criterion'' hyperparameter in GBM controls the metric used to measure the quality of a split in GBM, which seems to be essential in GBM.
Overall, it can be seen that only a small set of hyperparameters generally contribute towards the actual model performance.
However, the hyperparameters influential to the model might differ, and the set of hyperparameters generally differs significantly across model types.
Therefore, it is essential to obtain a rough estimate of hyperparameters' importance when deploying ML models on IoT detection tasks, such that we can account for the model's sensitivity to hyperparameters and tune their corresponding value when necessary.

\subsubsection{Visualising Parametric Relationship}

Continuing from the previous parametric study, we further study the relationship between each hyperparameter with respect to the model performance on its final F1 score.
\Cref{fig:parameter-relationship} illustrates a set of subplots that plots the value of each hyperparameter against their objective values.
Scatters with the same hue of colour refer to the same trial, and the scatters visualise the attainable objective values for various sets of hyperparameters and their distribution.

Some subplots illustrate a clear relationship between the choice of hyperparameters, while some hyperparameters do not indicate any apparent effect on the objective value.
For example, most hyperparameters for GBM in~\cref{fig:parameter-relationship:gbm} do not appear to contain any clear indications of best value, with the exception for ``Criterion'', where the Friedman Mean Square Error (MSE)~\citep{friedman2001_GreeFunc} is superior to the traditional MSE.
In contrast, hyperparameters for Bagging in~\cref{fig:parameter-relationship:bagging} exhibit clear trends.
On the one hand, a higher value of ``Max Features'' (the maximum number of features used to train sub-estimators) seems to provide a better objective value; on the other hand, a lower value of ``Max Samples'' (the maximum subset of the dataset used to sample for training sub-estimators) seems to yield better results.
Moreover, most hyperparameters for XGB appear to consistently score high objective value, with some exceptional outliers that occur when both ``Min Child Weight'' and ``N Estimators'' are close to zero.

It should be noted that the plots in~\cref{fig:optimisation-history} illustrates the complex interaction in-between hyperparameters; therefore, it is difficult to provide a definitive conclusive decision on the effect of a single hyperparameter.
However, the general trend of a hyperparameter can illustrate the model's sensitivity towards such a hyperparameter.
Moreover, the use of Bayesian Optimisation techniques like tree-structured Parzen estimator~\citep{bergstra2011_AlgoHype} helps to search for the set of hyperparameters that are best suited for our objectives.

\subsection{Most Influential Features for Detecting Cyber Attacks in IoT Devices}

This section summarises the numerous cybersecurity attacks studied in this paper and the type of features that are useful for identifying such an attack.
The degree of influential for each type of cyber attacks is computed using the absolute value of the pearson correlation coefficients.
The following summarises the top 3 most important features for detecting the studied IoT cyber attacks.

\begin{enumerate}[label=\roman*]
  \item \textbf{Distributed Denial of Service (DDoS)}: DDoS attack is a form of a malicious attempt to disrupt the normal traffic of some targeted server.
  These traffic flows are detected as part of a DDoS attack because of the number of flows directed to the same IP address.
  The essential network features that can identify whether a network package belongs to a DDoS attack are:
 (1) the originating or responding port number,
 (2) the maximum time between two packets sent in the backward direction, and
 (3) the number of packets with SYN / ACK flag being set.

  \item \textbf{Man-in-the-middle attack (MITM)}: MITM refers to an attacker being positioned between the IoT device and the communication endpoint to intercept and potentially alter data travelling between them.
  The network features that contribute primarily to identifying MITM attacks are:
  (1) the number of packets with ACK flag set,
  (2) the size (length) of packet in forward direction, and
  (3) the size of the package in the backward direction.

\item \textbf{Host Port Scan}: Port scanning is a method where attackers scope out their target environment by sending packets to specific ports on a host and using the responses to find vulnerabilities.
Notable network features that identify such attack includes:
(1) the number of packets with ACK flag set,
(2) the size of packet in backward direction, and
(3) the total number of bytes sent in the initial window in the backward direction.

\item \textbf{Mirai---HTTP Flooding / ACK Flooding}: \emph{Mirai} is a malware that exploits security holes in IoT devices and attempts to harness the collective power of millions of IoT devices into botnets for launching distributed attacks.
For Mirai-infected devices that are performing HTTP Flooding attacks (a type of DDoS), the following network features are most helpful in identifying them:
(1) the size of packet in forward direction,
(2) the size of packet in backward direction, and
(3) the total number of bytes sent in the initial window in the backward direction.
The same goes for ACK flooding attack.

\item \textbf{Mirai---UDP Flooding}: UDP flooding is a type of DDoS attack where a large number of User Datagram Protocol (UDP) packets are sent to a targeted server to overwhelm that device’s ability to process and respond.
The identifiable network features include:
  (1) the number of packets with ACK flag set,
  (2) the count of packets with at least 1 byte of TCP data payload in the forward direction, and
  (3) the number of backward packets per second.

\item \textbf{Mirai---Host brute force}: Brute force attacks executed by Mirai-infected devices attempt to gain access to a site or server by systematically trying every possible combination.
The typical preventative approach of blocking brute force attacks by locking out IP addresses would be less effective because the Mirai botnet is a distributed attack.
The important identifiable network features include:
    (1) the number of packets with ACK / SYN flag set,
    (2) the size of packet in backward direction, and
    (3) the number of flow bytes per second.
\end{enumerate}

The summarised network features insights on popular cyber security attacks are useful for IoT device designers and network security experts to design an integrated system that can monitor and potentially identify and prevents IoT devices from being hijacked by the cyber threats.

\section{Conclusion}

This paper presents a comprehensive study on applying ensemble machine learning methodologies for detecting cybersecurity attacks in an IoT environment.
A wide range of traditional and ensemble machine learning models is analysed, along with a Bayesian Optimisation framework and analysis of the sensitivity of the choice of hyperparameters on model performance.
This study highlighted the set of influential configurations on the popular models and the optimisation approach of automatic tunning of such hyperparameters with Bayesian Optimisation.
In addition, we evaluate the various models against a wide range of IoT anomaly datasets to evaluate each model's robustness.
Empirical evidence suggests that Tree-based boosting methods like GBM and XGB can consistently achieve high accuracy and recall.
Moreover, the parametric study on hyperparameters illuminates their importance and effects on the final model performance.
Most machine learning models contain a large set of hyperparameters, but only a small set exhibit a strong influence on the model performance.
Therefore, such hyperparameters should either be derived from best practice or by performing the present parametric study to determine important hyperparameters.

Cybersecurity, IoT devices and related technologies are the current focus in many multi-discipline domains, for example, electronics manufacturers on designing smart devices, network security analysts on improving the current network protocol, and data scientists on better utilising the vast amount of available data.
In particular, people rely heavily on IoT devices and services due to the wide range of applications in our daily lives.
A possible direction for future studies is testing more variety of networks protocol and device types as they can provide a more diverse set of representations for the model to extract more meaningful features.

\bibliography{references}      %

\end{document}